\documentclass[11pt,a4paper]{article}
\usepackage[hyperref]{naaclhlt2018}
\usepackage{times}
\usepackage{latexsym}
\usepackage{graphicx}
\graphicspath{ {image/} }
\usepackage{url}
\usepackage{color}
\usepackage{amsmath}

\DeclareMathOperator*{\argmin}{argmin}

\aclfinalcopy 

\title{UMDuluth-CS8761 at SemEval-2018 Task 9:  
Hypernym Discovery using Hearst Patterns, Co-occurrence frequencies and Word Embeddings}

\author{Arshia Z. Hassan \and Manikya S. Vallabhajosyula \and Ted Pedersen\\
        Department of Computer Science\\
        University of Minnesota\\
        Duluth, MN 55812 USA \\
  {\tt \{hassa418,valla045,tpederse\}@d.umn.edu} \\
  {\tt https://github.com/manikyaswathi/SemEval2018HypernymDiscovery}
}

\date{}

\begin{document}
\maketitle
\begin{abstract}
Hypernym Discovery is the task of identifying potential hypernyms for a given term. 
A hypernym is a more generalized word that is super-ordinate to more specific words. 
This paper explores several approaches that rely on co-occurrence frequencies 
of word pairs, Hearst Patterns based on regular expressions, and word embeddings created from 
the UMBC corpus. Our system Babbage participated in Subtask 1A for English and placed
6th of 19 systems when identifying concept hypernyms, and 12th of 18 systems for entity 
hypernyms.
\end{abstract}

\section{Introduction}

\begin{figure}[h]
    \centering
    \includegraphics[width=.25\textwidth]{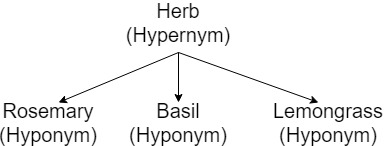}
    \caption{Hypernym-hyponym example}
    \label{fig:Hypernym-hyponym example}
\end{figure}

Hypernym-hyponym pairs exhibit an \textit{is-a} relationship where a hypernym is a 
generalization of a hyponym. The objective of SemEval--2018 Task 9 \cite{task9} is to generate a 
ranked list of hypernyms when given an input hyponym and a vocabulary of candidate hypernyms.
For example, the input hyponym \emph{lemongrass} could yield the hypernyms [\emph{grass, oil plant, herb}], where \emph{herb} would
be the best candidate. This scenario is illustrated in Figure \ref{fig:Hypernym-hyponym example}, where the three leaf nodes are
hyponyms and the root is a hypernym.

Note that hypernym discovery is distinct from hypernym detection,  
where the problem is to detect if a hyponym-hypernym relationship exists between a given pair, 
such as \emph{lemongrass-grass}. 

In our first module, we retrieve candidate hypernyms for an input term using a paragraph-length 
context-window and calculate their co-occurrence frequencies, which is later used for ranking the 
candidates. Our second module uses Hearst Patterns \cite{Hearst:1992:AAH:992133.992154} to 
extract hyponym-hypernym pairs and ranks candidate hypernyms based on co-occurrence frequency of the pairs. 
Our final module employs word-embeddings created using word2vec \cite{DBLP:journals/corr/abs-1301-3781}.
This paper continues with a more detailed discussion of each module, and then a review of our results.

\section{Implementation}

Babbage begins by pre-processing (\ref{Pre-processing}) the UMBC Corpus (\ref{corpus}) and 
extracting candidate hypernyms using four different strategies (\ref{extract}).
The first and second module calculates the co-occurrence frequencies between 
the input term and words in context using the pre-processed UMBC Corpus and the Hearst Pattern set 
extracted from the UMBC Corpus. The third module uses the IS-A Hearst Pattern set extracted from UMBC 
Corpus to obtain hypernyms. The final module constructs a word embedding over the UMBC corpus 
and uses a distance measure to fetch candidate hypernyms for a given input term.

\subsection{UMBC Corpus}\label{corpus}

Our training corpus is the University of Maryland, Baltimore County (UMBC) WebBase Corpus  
\cite{UMBC_EBIQUITY_CORE_Semantic_Textual_Similarity_Systems}. 
It contains 3 billion words from paragraphs obtained from more than 
100 million web pages over various domains. We use the 28 GB tokenized version of UMBC corpus 
which is part-of-speech tagged and divided among 408 files. 
There is also a vocabulary file with 218,755 unigram, bigram and trigram hypernym terms 
provided by task organizers. This file defines the set of possible candidate hypernyms.

\subsection{Architecture of System Babbage}
The following are the steps involved in constructing our system:
\begin{enumerate}

\item \textbf{Pre-processing the input text corpus [\ref{Pre-processing}]:} Corpora obtained in this stage include:

  \begin{enumerate}
     \item Normalize the input corpus and store as \textit{Normalized Corpus}
     \item Fetch Hearst Patterns (see Figure 2) from input corpus and store as \textit{Hearst Corpus}
     \item Fetch IS-A Pattern from the input corpus and store as \textit{IS-A Corpus}
     \item Creating the word-embedding matrix \textit{UMBC Embedding} using Normalized Corpus.
  \end{enumerate}

The Hearst Corpus and the IS-A Corpus patterns are extracted from the 
original text corpus which has been preprocessed to eliminate punctuation, prepositions, 
and conjunctions. All possible combinations of bigram and trigram noun phrases are 
retained in the Normalized Corpus. A Word-Embedding matrix is built over this Normalized Corpus.

  \item \textbf{Extracting candidate hypernyms [\ref{extract}]:}
  \begin{enumerate}
      \item \textbf{Co-occurrence frequencies from Normalized Corpus:} A co-occurrence map is built 
for the input terms with the words in the context of the input term and the 
frequency of their co-occurrence using the Normalized Corpus. Words with co-occurrence 
frequency higher than \textit{\textbf{5}} are listed as candidate hypernyms for an input term. 
This is considered the first module result.
      \item \textbf{Co-occurrence frequencies from Hearst Corpus:} A co-occurrence map similar to the previous 
step is built by using the Hearst Corpus. All the words which occur at least once in 
context of the input term in the Hearst Patterns are listed as candidate hypernyms for this term. 
This is considered the second module result.
      \item \textbf{Co-occurrence frequencies from IS-A Corpus:} All the words which occur at least once 
in the context of the input term in the IS-A Corpus are listed as candidate 
hypernyms for this term. If the input term is a concept and is a bigram or 
trigram term, then part of it is considered as a hypernym for that term. 
This is considered the third module result.
      \item \textbf{Applying word similarity to word embeddings:} A fixed distance value called 
\textit{\textbf{Phi}} \cite{fu2014learning} is used to extract words at this distance to the input term  in the UMBC Embedding. 
These words are listed as the candidate hypernyms for an input term.  
This is considered our final module result.
    \end{enumerate}
  \item \textbf{Merging results from various modules [\ref{ranking}]:} The order of merging these results is 
decided by the evaluation scores from these modules for training data. The same order is applied to the test data.
  
\end{enumerate}

\subsubsection{Pre-processing}\label{Pre-processing}

The task description states that our system should predict 
candidate hypernyms for an input word which is either a concept or an entity. 
Hence, the part-of-speech tag for all candidate hypernyms is \textit{noun}. 
This restricts our search space to words with \textit{noun} 
part-of-speech tag and bigram or trigram phrases with a \textit{noun} head word. 
Our system focuses on concepts, so we 
do not have any module specific for entities. To refine the input corpus as 
per these specifications, the input UMBC Corpus is processed through the following modules:

     \paragraph{Normalized Corpus:}\label{normP}
     The POS tagged input corpus is processed per paragraph. 
Each paragraph is converted to lower-case text. Then, bigram and trigram noun 
phrases from each paragraph are obtained using the POS tags given for 
each word. It is further filtered by removing punctuation marks and words 
with part-of-speech tags other than \textit{noun}, \textit{verb}, \textit{adverb} 
or \textit{adjective}. This filtered line is modified by appending it with bigram 
and trigram noun phrases obtained earlier.

\begin{figure*}[t]
        \centering
        \includegraphics[width=.78\textwidth]{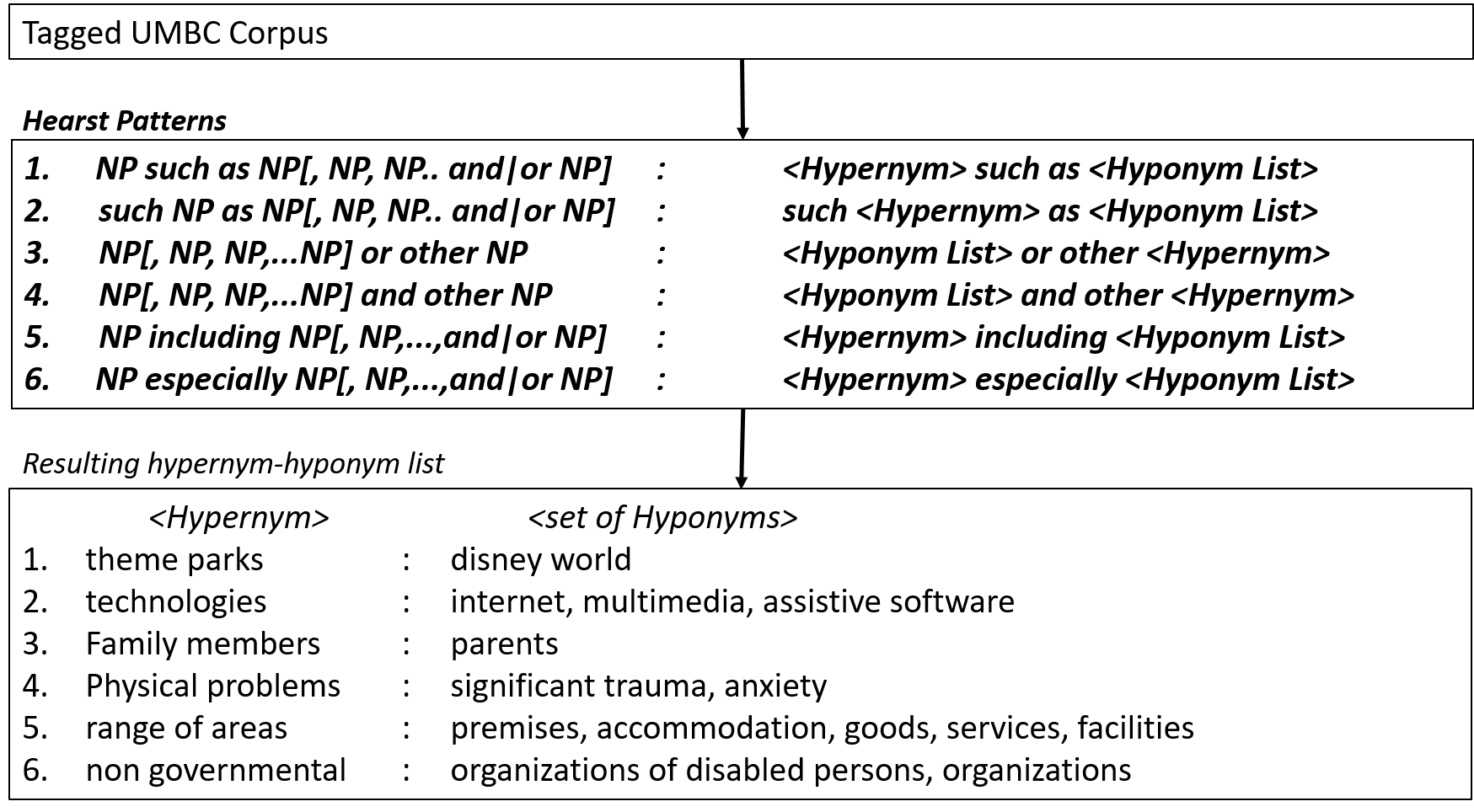}
        \caption{Creating Hearst Corpus using 6 Hearst Patterns}
        \label{fig:Obtaining Hearst patterns}
    \end{figure*}

     \paragraph{Hearst Corpus:}\label{hearstP}
     The original input paragraph is searched for the Hearst Patterns 
(shown in Figure \ref{fig:Obtaining Hearst patterns}) and all the possible 
matches are returned in the form of \textit{hypernym} : \textit{one or more hyponyms}. 
Figure \ref{fig:Obtaining Hearst patterns} shows the extraction of Hearst Patterns,  
where NP represents a noun-phrase where the head word is tagged as a \textit{noun},  
\textbf{\textit{the loved-ones} such as \textit{family and friends}} is a match for Hearst Patterns 
(from Figure \ref{fig:Obtaining Hearst patterns}) with noun phrases \textit{the loved-ones},  \textit{family} and \textit{friends}.

     \paragraph{IS-A Corpus:}\label{is-aP}
     A pattern which is not used in the construction of the Hearst Corpus is used here: 
\textbf{Hyponym Noun Phrase \textit{is (a $\vert$ an $\vert$ the)} Hypernym Noun Phrase}. Here the original input paragraph is searched against this pattern and all the possible matches 
are returned in the form of \textit{hyponym} : \textit{hypernym}. \textbf{\textit{a fennel} 
is a \textit{plant}} is a match for this pattern with noun phrases \textit{a fennel} and \textit{plant}.
     \paragraph{UMBC Embedding:}\label{embedP}
     A word embedding matrix is created over the Normalized Corpus using word2vec \cite{DBLP:journals/corr/abs-1301-3781}. 
The specifications of the model are as follows: 

     (a) \textbf{Model :} \textbf{\textit{Continuous Bag of Words (CBOW)}} - a term's embedding value is determined by 
its context words. The order of the words in the window size does not matter. 

(b) \textbf{Window Size :} \textit{\textbf{10}}. 
The context window size for a term which determines its vector value. 

(c) \textbf{Minimum Frequency Count :} \textbf{\textit{5}}. If the frequency of a word is less 
than this value, the word does not exist in the embedding. 

(d) \textbf{Embedding Dimension Size :} \textbf{\textit{300}}. The number of dimensions for the embedding matrix.

\subsubsection{Extracting candidate hypernyms:}\label{extract}

Once the UMBC corpus is pre-processed and the three required corpora and 
an embedding matrix are derived, candidate hypernyms are acquired by applying the below processes. 

    \paragraph{Co-occurrence frequency from Normalized Corpus:}\label{normC}
    With this module, we hypothesized that a hyponym and its possible 
hypernyms are more likely to co-occur within a context-window. 
The context window of a term is its own paragraph. We start by creating 
a map for all the input terms. If a normalized paragraph \ref{normP} contains any 
of the input terms, then all the words in the context are added to the map of this 
particular term which considers them to be hypernyms for this input hyponym term. 
Every time a hypernym-hyponym pair co-occurs in one line, their co-occurrence 
count is increased by one. Finally, the candidate hypernyms are ranked in 
descending order of their co-occurrence frequencies.

    \paragraph{Co-occurrence frequency from Hearst Corpus:}\label{hearstC}
    In the pre-processing step \ref{Pre-processing}, we extracted possible hypernym-hyponym mapping data using Hearst Patterns. 
Each line of the data is of the form \textit{hypernym : hyponym-1 , hyponym-2, … , hyponym-n}. 
In this module, we created a map where each \textit{hyponym} is a key 
mapped to \textit{hypernyms} occurring with that \textit{hyponym} and their co-occurrence frequencies. 
For example, values for keys \textit{hyponym-1, hyponym-2, … and hyponym-n} 
are updated with \textit{hypernym} and the frequencies are increased by 1. Finally the top 15 \textit{hyponyms} 
(based on frequencies) for each key are reported as the result hypernyms.
    
    \paragraph{Co-occurrence frequencies from IS-A Corpus:}\label{is-aC}
    This module uses hypernym-hyponym pairs from the IS-A Corpus \ref{Pre-processing} 
which are in the form \textit{hyponym : hypernym}. We use the same strategy as 
\textit{Co-occurrence frequency from Hearst Corpus} to obtain the result.
    
    \paragraph{Applying word similarity to word embeddings:}\label{phi-m}
    We need a general distance vector which represents a hypernym-hyponym 
distance in the UMBC Embedding. We use training data input term (x) and the 
gold data hypernyms (y) to calculate this distance(\( \Phi^* \)) which is calculated by:

\begin{equation}
       \Phi^* =  \underset{\Phi}{\argmin} \frac{1}{N}\sum_{(x, y)} \Phi \| x - y \|^2 
\end{equation}
    $\Phi$ is used to get candidate hypernyms from the UMBC word embedding matrix for the input terms (test data). 

\subsection{Merging results from various modules} \label{ranking}

For this task, our system is required to report the 15 most probable hypernyms for each input term. 
We have four modules each reporting their top 15 candidate hypernyms. By looking at the training scores 
of these modules, we merge the co-occurrence frequencies from IS-A corpus that have higher ranks followed by 
the co-occurrence frequencies from Normalized corpus and Hearst Pattern corpus. 
Results from word embedding module are given the lowest ranks. 

\section{Experimental Results and Discussion}

Output candidate hypernym lists are evaluated against gold hypernym lists using the 
following evaluation criteria : Mean Reciprocal Rank (MRR),  Mean Average Precision 
(MAP) and Precision At k (P@k), where k is 1, 3, 5 and 15. We ran our model against 
two sets of data training data and test data with 1500 input terms each. These
results are shown in Tables 1 and 2, where it can be clearly observed that our system
performs much better for concepts. However, the IS-A module seemed to fetch good candidates
for both entity and concept data. 

\begin{table}
\centering
\begin{tabular}{|c|c|c|c|c|c|}
 \hline
   & Cooc & Hearst & Phi & Is-A & Merged\\
  \hline
 MRR & .103 & .020 & .025 & .165 & .188\\
 MAP & .050 &  .008 & .012 & .071 & .080\\
 P@1 & .061 & .013 & .012 & .140 & .152\\
 P@3 & .055 & .008 & .012 & .076 & .087\\
 P@5 & .048 & .008 & .012 & .066 & .075\\
 P@15 & .047 & .007 & .011 & .062 & .070\\
  \hline
\end{tabular}
\caption{\label{Test Data Concept}
Test Data 1A English - Concept Scores
  }
\end{table}

\begin{table}
\centering
\begin{tabular}{|c|c|c|c|c|c|}
 \hline
   & Cooc & Hearst & Phi & Is-A & Merged\\
  \hline
 MRR & .000 & .000 & .008 & .090 & .099\\
 MAP & .000 &  .000 & .003 & .036 & .037\\
 P@1 & .000 & .000 & .004 & .069 & .081\\
 P@3 & .000 & .000 & .003 & .041 & .045\\
 P@5 & .000 & .000 & .003 & .035 & .036\\
 P@15 & .000 & .000 & .003 & .031 & .030\\
  \hline
\end{tabular}
\caption{\label{Test Data Entity}
Test Data 1A English - Entity Scores
  }
\end{table}

The gold data provided with the task does not always consider all 
possible word senses or domains of an input term. As a result, 
we observed numerous candidate hypernyms that seem to be plausible solutions 
that are not considered correct when compared to the gold data. 

For example, the input concept \textit{navigator} has gold standard hypernyms of 
\textit{[military branch, explorer, military machine, 
travel, adventurer, seaman]}. Our system finds candidate hypernyms \textit{[browser, web browser, website, application]}. 
We also noticed that due to our normalization decisions (i.e., using all lower-case characters) and the contents of the 
corpus, Babbage performs poorly in some cases. For example, the gold hypernyms for input entity 
\textit{Hurricane} are \textit{[video game, software program, computer program]} but our system 
produced \textit{[storm, windstorm, typhoon, tornado, cyclone]}. 
Clearly, our system did not differentiate between the named entity \textit{Hurricane} and the common 
noun \textit{hurricane} while training the word--embedding models. 


On the positive side, our system produced promising results in some cases. 
Hyponym \textit{liberalism} produced \textit{[theory, philosophy, economic policy]} which is 
very similar to the gold data \textit{[economic theory, theory]}. It also correctly generated the 
hyponym \textit{person} for hypernyms such as collector, moderator, director, senior, and reporter. 
For input \textit{reporter} it produced \textit{[writer, person]} which matches the gold hypernym set.

\section*{Acknowledgments}

This project was carried out as a part of CS 8761, Natural Language Processing, a graduate level class offered 
in Fall 2017 at the University of Minnesota, Duluth by Dr. Ted Pedersen. All authors of this paper have 
contributed equally and are listed in alphabetical order by first name.

\bibliography{babbage_paper.bib}
\bibliographystyle{acl_natbib}

\end{document}